\documentclass[final]{cvpr}

\usepackage{times}
\usepackage{epsfig}
\usepackage{graphicx}
\usepackage{amsmath}
\usepackage{amssymb}
\usepackage{soul,color}
\usepackage{amsmath}
\usepackage[ruled,lined]{algorithm2e}
\usepackage{tabularx}
\usepackage{booktabs}

\usepackage{epstopdf}
\usepackage{color, colortbl}
\usepackage{makecell}
\usepackage{multirow}

\usepackage{bbold}

\usepackage[pagebackref=true,breaklinks=true,colorlinks,bookmarks=false]{hyperref}

\def\cvprPaperID{} 
\def\confYear{CVPR 2021}

\begin{document}

\title{TinyDefectNet: Highly Compact Deep Neural Network Architecture for High-Throughput Manufacturing Visual Quality Inspection}
\author{Mohammad Javad Shafiee$^{1,2}$, Mahmoud Famouri$^{2}$, Gautam Bathla$^{2}$, Francis Li$^{2}$,  Alexander Wong$^{1,2}$ \\
$^1$University of Waterloo, Waterloo, Ontario, Canada\\
$^2$DarwinAI,  Waterloo, Ontario, Canada \\
mjshafiee@uwaterloo.ca  \\
\{mahmoud, gautam.bathla, francis\}@darwinai.ca\\
alexander.wong@uwaterloo.ca
}

\maketitle
\begin{abstract}
\vspace{-0.35cm}
    A critical aspect in the manufacturing process is the visual quality inspection of manufactured components for defects and flaws.  Human-only visual inspection can be very time-consuming and laborious, and is a significant bottleneck especially for high-throughput manufacturing scenarios.  Given significant advances in the field of deep learning, automated visual quality inspection can lead to highly efficient and reliable detection of defects and flaws during the manufacturing process. However, deep learning-driven visual inspection methods often necessitate significant computational resources, thus limiting throughput and act as a bottleneck to widespread adoption for enabling smart factories. In this study, we investigated the utilization of a machine-driven design exploration approach to create TinyDefectNet, a highly compact deep convolutional network architecture tailored for high-throughput manufacturing visual quality inspection.  TinyDefectNet comprises of just $\sim$427K parameters and has a computational complexity of $\sim$97M FLOPs, yet achieving a detection accuracy of a state-of-the-art architecture for the task of surface defect detection on the NEU defect benchmark dataset.  As such, TinyDefectNet can achieve the same level of detection performance at 52$\times$ lower architectural complexity and 11$\times$ lower computational complexity. Furthermore, TinyDefectNet was deployed on an AMD EPYC 7R32, and achieved 7.6$\times$ faster throughput using the native Tensorflow environment and 9$\times$ faster throughput using AMD ZenDNN accelerator library.  Finally, explainability-driven performance validation strategy was conducted to ensure correct decision-making behaviour was exhibited by TinyDefectNet to improve trust in its usage by operators and inspectors.
\end{abstract}

\vspace{-0.45cm}
\begin{figure}[t]
    \centering
    \includegraphics[width=8cm]{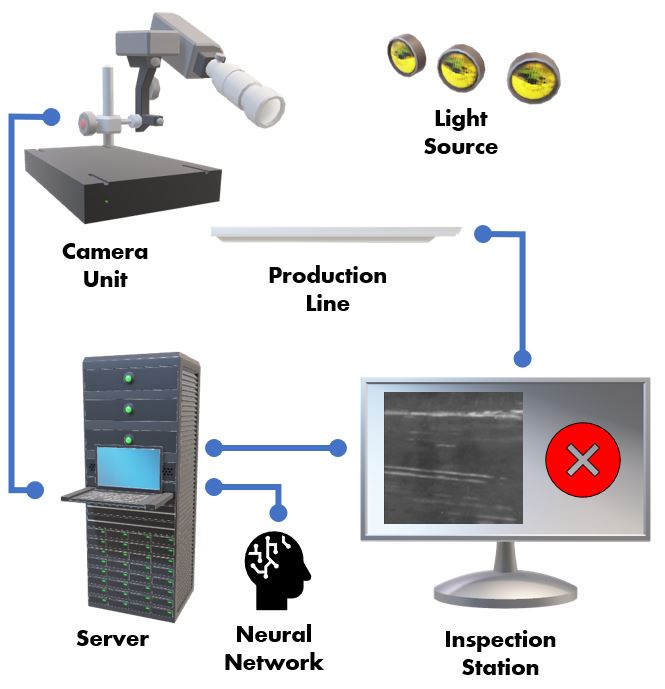}
    \caption{Surface defect visual inspection system. Surface defect inspection is an example of a manufacturing visual quality inspection application that can greatly benefit from machine learning systems.}
    \label{fig:system}
\end{figure}
\section{Introduction}
The promises of the new era for machine learning have motivated different industries to augment and enhance the capabilities and efficiency of their existing workforce with artificial intelligence (AI) applications that automate highly-repetitive and time-consuming processes. In particular, the advances in deep learning have led to promising results in different applications ranging from computer vision tasks~\cite{he2016deep,robinson2020learning} such as image classification~\cite{he2016deep} and video object segmentation~\cite{robinson2020learning} to natural language processing tasks~\cite{devlin2021bert} such as language translation~\cite{ranathunga2021neural} and question-answering~\cite{rongali2020exploring}. However, much of success in adopting deep learning in real-world applications have been in newer technology sectors such as e-commerce and social media where the tasks are inherently automation-friendly without human intervention under a controlled environment (e.g., product/content recommendation), with very limited adoption in traditional industrial sectors such as manufacturing where the tasks are currently done manually by human agents in unconstrained environments.

As an example, the visual quality inspection of manufactured components to identify defects and flaws is a critically important task in manufacturing that is very laborious and time-consuming (See Figure~\ref{fig:system} for example of a surface defect visual inspection system).  As such, automating this process would result in significant cost savings and significantly improved production efficiencies.  However, there are a number of key challenges that make it very difficult to adopt deep learning for visual quality inspection in a real-world production setting, including: i) small data problem: the availability of annotated data is limited and thus makes building highly accurate deep learning models a challenge, and ii) highly constrained operational requirements: designing high-performance deep learning models that satisfy very constrained operational requirements with regards to inference speed and size is very challenging especially in high-throughput manufacturing scenarios.

A common practice to the development of custom deep learning models for new industrial applications is to take advantage of off-the-shelf generic models published in research literature such as ResNet~\cite{he2016deep} and MobileNet~\cite{sandler2018mobilenetv2} architectures and apply transfer learning~\cite{zhuang2020comprehensive} to learn a new task given the available training data samples. While this approach enables rapid prototyping of models to understand feasibility of the task at hand, such generic off-the-shelf models are not tailored for specific industrial tasks and are often unable to meet operational requirements related to speed and size for real-world industrial deployment.  As such, it makes it very challenging to adopt off-the-shelf generic models for tackling visual quality inspection applications under real-world manufacturing scenarios.

A very promising strategy for the creation of highly-customized deep learning models tailored for manufacturing visual quality inspection applications is machine-driven design exploration, where the goal is to automatically identify deep neural network architecture designs based on operational requirements.  One path towards machine-driven design exploration is neural architecture search (NAS)~\cite{liu2018darts,tan2019mnasnet,ren2021comprehensive}, where the problem is formulated as a large-scale, high-dimensional search problem and solved using algorithms such as reinforcement learning and evolutionary algorithms. While they have shown promising results in designing new deep neural network architectures~\cite{ren2021comprehensive}, they are very computationally intensive and require very large-scale computing resources over long search times~\cite{ren2021comprehensive}.   More recently, another path towards machine-driven design exploration is the concept of generative synthesis~\cite{wong2018ferminets}, where the problem is formulated as a constrained optimization problem and the approximate solution is found in an iterative fashion using a generator-inquisitor pair.  This approach has been demonstrated to successfully generate deep neural network architectures tailored for different types of tasks across different fields and applications~\cite{wong2020tinyspeech,abbasi2021outliernets,wang2020covid}.

In this study, we explore the efficacy of machine-driven design exploration for the design of deep neural network architectures tailored for high-throughput manufacturing visual quality inspection.  Leveraging this strategy and a set of operational requirements, we introduce TinyDefectNet, a highly compact deep convolutional network architecture design automatically tailored around visual surface quality inspection. Furthermore, we evaluate the capability of the ZenDNN accelerator library for AMD processors in further reducing the run-time latency of the generated TinyDefectNet for high-throughput inspection.

 \begin{figure*}
    \centering
    \includegraphics[width=18cm]{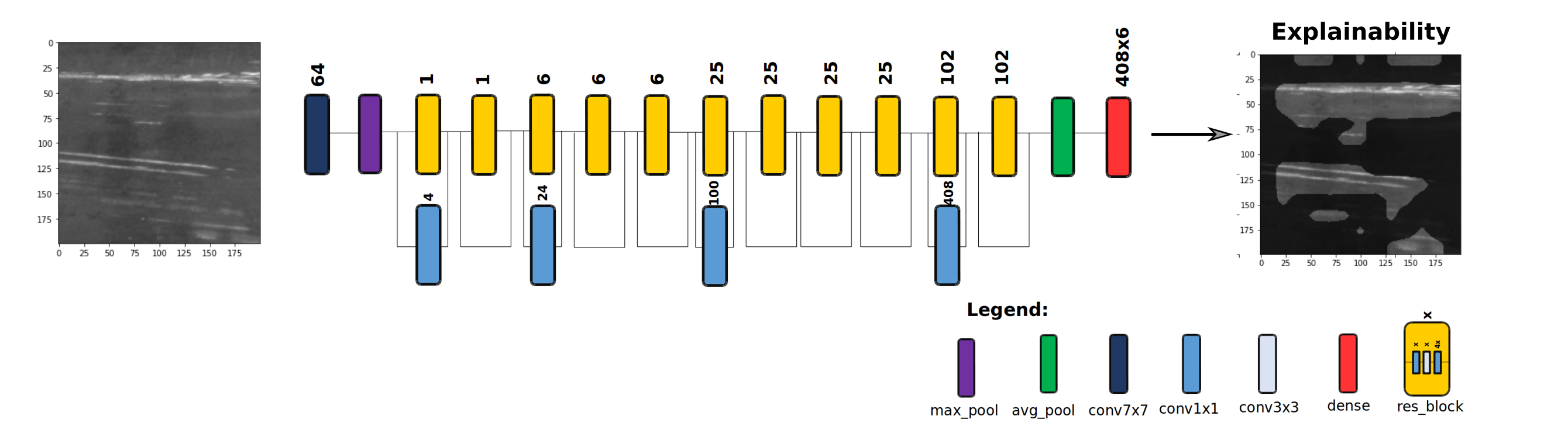}
    \caption{The network architecture of the proposed TinyDefectNet. The proposed architecture design, as produced via machine-driven design exploration, possesses a shallow architecture with heterogeneous, lightweight micro- and macro-architecture traits that are well-suited for high-throughput inspection scenarios.  We conduct explainability-driven performance validation on TinyDefectNet (the input surface image (left) and the corresponding explanation (right) that highlights the critical factors behind the decision-making process) to ensure correct decision-making behaviour and improve trust in its usage.}
    \label{fig:arch}
\end{figure*}

\section{Methodology}
The macro- and micro-architectures of the proposed TinyDefectNet are designed by leveraging the concept of generative synthesis for the purpose of machine-driven design exploration.  The concept of generative synthesis revolves around the formulation of the design exploration problem as a constrained optimization problem.   More specifically, we wish to find an optimal generator $G^\star(\cdot)$ given a set of seeds $S$ which can generate networks architectures $\{\mathcal{N}_s|s \in  S\}$ that maximize a universal performance
function $U$, with constraints defined by a predefined set of  operational requirements formulated via an indicator function $\mathbb{1}_g(\cdot)$,

\begin{equation}
    G^{\star} = \underset{G^{'}}{\max} U \Big(G(s)\Big) \;\;\; \text{s.t.} \;\;\; \mathbb{1}_g(G(s)) = 1 \;\;\; \forall s \in S.
\end{equation}

\begin{table*}[h]
    \centering
    \caption{Quantitative results of the proposed TinyDefectNet architecture compared to the off-the-shelf ResNet50 architecture. It can be observed that the proposed architecture is 56$\times$ smaller with 11$\times$ lower computational complexity in terms the number of FLOPs. This results to a deep learning model which is 7.6$\times$ faster compared to ResNet-50 on an AMD EYPC 7R32 processor.}
    \begin{tabular}{l||c|r|r|c}
        \bf Model &\bf ACC & \bf  Parameters &\bf  FLOPs\;\;\;\;\; &\bf  Inference Speed (s)  \\\hline\hline
        \bf ResNet-50& 98\%& 24,136,710 & 1,115,962,374 & 0.01881 \\
        \bf  TinyDefectNet&98\% &427,776&97,263,435 &0.00247\\\hline
        \bf Improvement & -- &56$\times$\;\;\;\;\;\;& 11$\times$ \;\;\;\;\;\;& 7.6$\times$
    \end{tabular}

    \label{tab:res}
\end{table*}

Finding $G^\star(\cdot)$ in a direct manner is computationally infeasible. As such, we find the approximate solution to $G^\star(\cdot)$ through an iterative optimization process, where in each step the previous generator solution $\bar{G}(\cdot)$ is evaluated by an inquisitor $I$ via its newly generated architectures $\mathcal{N}_s$, and this evaluation is used to produce a new generator solution.  The initiation of this iterative optimization process is conducted based on a prototype, $U$, and $\mathbb{1}_g(\cdot)$.

In this study, we define a residual design prototype $\phi$ based on the principles proposed in~\cite{he2016deep}, define $U$ based on~\cite{wong2019netscore}, and define the indicator function $\mathbb{1}_g (\cdot)$ with the following operational constraint: number of floating-point operations (FLOPs) is within 5\% of 100M FLOPs to account for high-throughput manufacturing visual inspection scenarios.

The network architecture of TinyDefectNet is demonstrated in Figure~\ref{fig:arch}, and there are two key observations worth highlighting in more detail. First, it can be observed that the micro- and macro-architecture designs of the proposed TinyDefectNet possess heterogeneous, lightweight characteristics that strike a strong balance between representational capacity and model efficiency.  Second, it can also be observed that TinyDefectNet has a shallower macro-architecture design to facilitate for low latency, making it well-suited for high-throughput inspection scenarios. These architectural traits highlight the efficacy of leveraging a machine-driven design exploration strategy in the creation of high-performance customized deep neural network architectures tailored around task and operational requirements needed for a given application.

\section{Results \& Discussion}
The generated TinyDefectNet architecture is evaluated using the NEU-Det benchmark dataset~\cite{NEU-Det} for surface defect detection, with the performance of the off-the-shelf ResNet-50 architecture~\cite{he2016deep} also evaluated for comparison purposes.  Both architectures were implemented in Keras with a Tensorflow-backend.

\begin{figure}
    \centering
\begin{tabular}{ccc}
     \includegraphics[width=2.5cm]{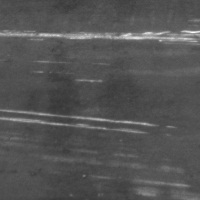}&\includegraphics[width=2.5cm]{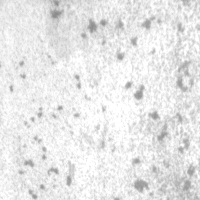}  &\includegraphics[width=2.5cm]{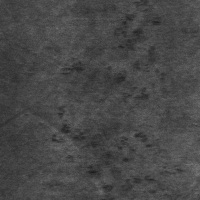} \\
     (a) & (b) & (c)\\
     ~\\
     \includegraphics[width=2.5cm]{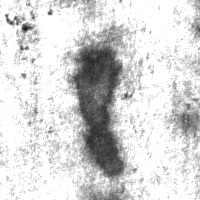}& \includegraphics[width=2.5cm]{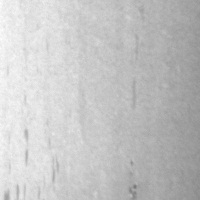} & \includegraphics[width=2.5cm]{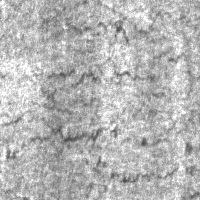}\\
     (d) & (e) & (f)
\end{tabular}
    \caption{Examples of different surface defect types: (a) Scratches, (b) Pitted surface, (c) Rolled in scale, (d) Patches, (e) Inclusion and (f) Crazing.}
    \label{fig:samples}
\end{figure}

The NEU-Det~\cite{NEU-Det} benchmark dataset used in this study is a metallic surface defect dataset comprising of 6  different defects including rolled-in scale (RS), patches (Pa), crazing (Cr), pitted surface (PS), inclusion (In) and scratches (Sc) (See Figure~\ref{fig:samples} for examples of surface defects). The dataset contains  1,800 grayscale images with equal number of samples from each classes; from 300 images of each class, 240 images are assigned for training and 60 images for testing. The input image size to the evaluated deep learning models is $200 \times 200$.

\subsection{Performance Analysis}
Table~\ref{tab:res} shows the quantitative performance results of the proposed TinyDefectNet. It can be observed that the proposed TinyDefectNet network architecture comprises of only $\sim$427K parameters, which is 56$\times$ smaller compared to an off-the-shelf ResNet-50 architecture. Furthermore, in terms of computational complexity, the proposed TinyDefectNet architecture requires only $\sim$97M FLOPs to process the input data compared to 1.1B FLOPs required by the ResNet-50 architecture with the same input size, and thus requires 11$\times$ fewer number of FLOPs in comparison. This high efficiency is highly desirable especially given that the proposed TinyDefectNet architecture performs with the same accuracy as ResNet50 architecture.

To further evaluate the efficiency of the proposed TinyDefectNet model, its running time latency is examined (at a batch size of 1024) on an AMD CPU (in this study, an AMD EPYC 7R32 processor) and compared with the ResNet-50 architecture. As shown in Table~\ref{tab:res}, the proposed TinyDefectNet architecture can process an input image in 2.5~ms, which is a 7.6$\times$ speed gains when compared to the ResNet-50 architecture which needs 19~ms to process the same image.  The significant speed gains, complexity reductions demonstrated by the proposed TinyDefectNet over off-the-shelf architectures while achieving high accuracy makes it highly suited for high-throughput manufacturing inspection scenarios and speaks to the efficacy of leveraging a machine-driven design exploration strategy for producing highly tailored deep neural network architectures catered specifically for industrial tasks and applications.
\begin{table}[]
    \centering
    \caption{The optimal environment variables to further improve the performance of the  proposed TinyDefectNet via ZenDNN accelerator library. }
    \begin{tabular}{l|c}
        \bf Flag & \bf Value\\ \hline\hline
        ZENDNN PRIMITIVE CACHE CAPACITY & 4  \\
        ZENDNN BLOCKED FORMAT & 0 \\
        ZENDNN MEMPOOL ENABLE& 1\\
        ZENDNN TENSOR POOL LIMIT& 1\\
        ZENDNN TF CONV ADD FUSION {SAFE} &0 \\
        OMP NUM THREADS& 8\\
        GOMP CPU AFFINITY&0-7 \\
        & \\
    \end{tabular}

    \label{tab:zendnn}
\end{table}

\begin{figure}
    \centering
    \includegraphics[width = 8cm]{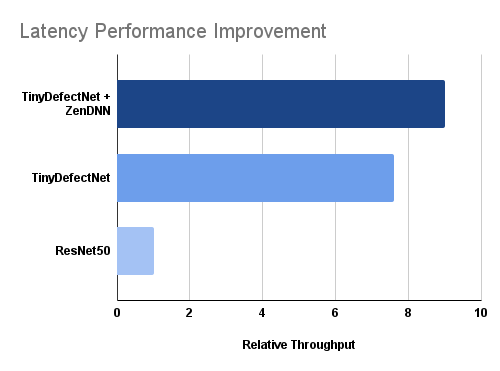}
    \caption{Performance comparison for the proposed TinyDefectNet with ZenDNN run-time accelerator, TinyDefectNet in native environment, and ResNet-50 in native environment. As seen, TinyDefectNet with ZenDNN can perform 9$\times$ faster compared to a ResNet-50 architecture in native environment.  }
    \label{fig:latency}
\end{figure}
\subsection{ZenDNN Acceleration}
In this section we analyze the impact of ZenDNN accelerator library for AMD processors on the performance of the proposed TinyDefectNet.  ZenDNN~\cite{ZenDNN} is a run-time accelerator library that is easy to use as it does not require model re-compilation or conversion and can be applied to different models for different tasks.

To further improve the running time latency of the proposed TinyDefectNet to improve throughput performance, we perform inference of TinyDefectNet within the ZenDNN environment. The optimal parameters for the ZenDNN variables are reported in Table~\ref{tab:zendnn}.  As shown in Figure~\ref{fig:latency}, leveraging the ZenDNN accelerator improves the performance of the proposed TinyDefectNet on an AMD CPU by 1.2$\times$ (i.e., $\sim$2ms).  As such, the proposed TinyDefectNet when running with ZenDNN can perform inference  9$\times$ faster when compared to the ResNet-50 architecture in a native Tensorflow environment.

\subsection{Explainability-driven Performance Validation}

The proposed TinyDefectNet was audited using an explainability-driven performance validation strategy to gain deeper insights into its decision-making behaviour when conducting visual quality inspection and ensure that its decisions are driven by relevant visual indicators associated with surface defects.  In particular, we leverage the quantitative explainability strategy proposed in~\cite{lin2019explanations}, which has been shown to provide good quantitative explanations that better reflect decision-making processes than other approaches in literature, and has been shown to be effective at not only model auditing~\cite{wang2020covid} but also identifying hidden data issues~\cite{wong2021insights}.  An example of an input surface image and the corresponding quantitative explanation corresponding explanation are shown in Figure~\ref{fig:arch}. It can be observed based on the quantitative explanation that TinyDefectNet correctly decided that this particular surface image exhibit scratch defects by correctly leveraging the different scratch defects found on the surface during its decision-making process.  In addition to ensuring correct model behaviour, conducting this explainability-driven performance validation process helps to improve trust in its deployment and usage by human operators and inspectors.

\section{Conclusion}
In this study we introduced TinyDefectNet, a highly effective and efficient deep neural network architecture for the high-throughput manufacturing visual quality inspection. We take advantage of generative synthesis for machine-driven design exploration to design micro- and macro-architectures of the proposed TinyDefectNet architecture in an automated manner. Experimental results show that the proposed model is highly efficient which can perform 9$\times$ faster when it runs with ZenDNN accelerator on an AMD CPU compared to an off-the-shelf ResNet-50 architecture in a native Tensorflow environment. This is very desirable especially since the proposed TinyDefectNet performs with 98\% accuracy at the same level as ResNet-50 architecture.  Explainability-driven performance validation strategy was conducted in this study to ensure correct decision-making behaviour was exhibited by TinyDefectNet to improve trust in its deployment and usage.  Future work involves  leveraging this machine-driven design exploration strategy for producing high-performing, highly efficient deep neural network architectures for other critical manufacturing tasks such as component localization and defect segmentation.

\bibliographystyle{IEEEtran}
\bibliography{references}
\end{document}